\documentclass[letterpaper, 10 pt, conference]{ieeeconf}  

\IEEEoverridecommandlockouts                              

\usepackage{graphics} 
\usepackage{subcaption}
\usepackage{multirow}
\usepackage{epsfig} 
\usepackage{mathptmx} 
\usepackage{times} 
\usepackage{amsmath,amssymb,amsfonts,mathrsfs} 
\usepackage{txfonts} 
\usepackage{multirow}

\usepackage{todonotes}
\usepackage{soul}
\definecolor{smoothgreen}{rgb}{0.7,1,0.7}
\sethlcolor{smoothgreen}

\title{\LARGE \bf
Orientation-Discriminative Feature Representation for \\ Decentralized Pedestrian Tracking
}

\author{Vikram Shree$^{1}$, Carlos Diaz-Ruiz$^{1}$, Chang Liu$^{1}$, Bharath Hariharan$^{2}$ and Mark Campbell$^{1}$
\thanks{*This work is supported by ONR BRC grant N00014-17-1-2175.}
\thanks{$^{1}$Vikram Shree, Carlos Diaz-Ruiz, Chang Liu, and Mark Campbell are with Sibley School of Mechanical and Aerospace, Cornell University, USA
        {\tt\small \{vs476, cad297, cl775, mc288\}@cornell.edu}}%
\thanks{$^{2}$Bharath Hariharan is with the Department of Computer Science, Cornell University, USA
        {\tt\small {bharathh}@cs.cornell.edu}}%
}

\begin{document}

\maketitle
\thispagestyle{empty}
\pagestyle{empty}

\begin{abstract}
This paper focuses on the problem of decentralized pedestrian tracking using a sensor network. Traditional works on pedestrian tracking usually use a centralized framework, which becomes less practical for robotic applications due to limited communication bandwidth. Our paper proposes a communication-efficient, orientation-discriminative feature representation to characterize pedestrian appearance information, that can be shared among sensors. Building upon that representation, our work develops a cross-sensor track association approach to achieve decentralized tracking. Extensive evaluations are conducted on publicly available datasets and results show that our proposed approach leads to improved performance in multi-sensor tracking.


\end{abstract}

\section{INTRODUCTION}
Using sensor networks and sensor-equipped mobile robots to visually track people in crowded outdoor environments over time, known as the Multi-Target Multi-Sensor Tracking (MTMST), has gained great interests in recent years due to its wide applications in visual surveillance, environmental monitoring, anomaly detection, and search and rescue. In practice, multiple sensors are usually deployed to achieve wide coverage. As a consequence, large changes of viewpoints, varying illumination conditions, and extended duration of missing or occluded detections are common \cite{javed2003tracking} and make recognizing and tracking people across sensors a challenging problem.

\begin{figure}
    \centering
    \includegraphics[width=.48\textwidth]{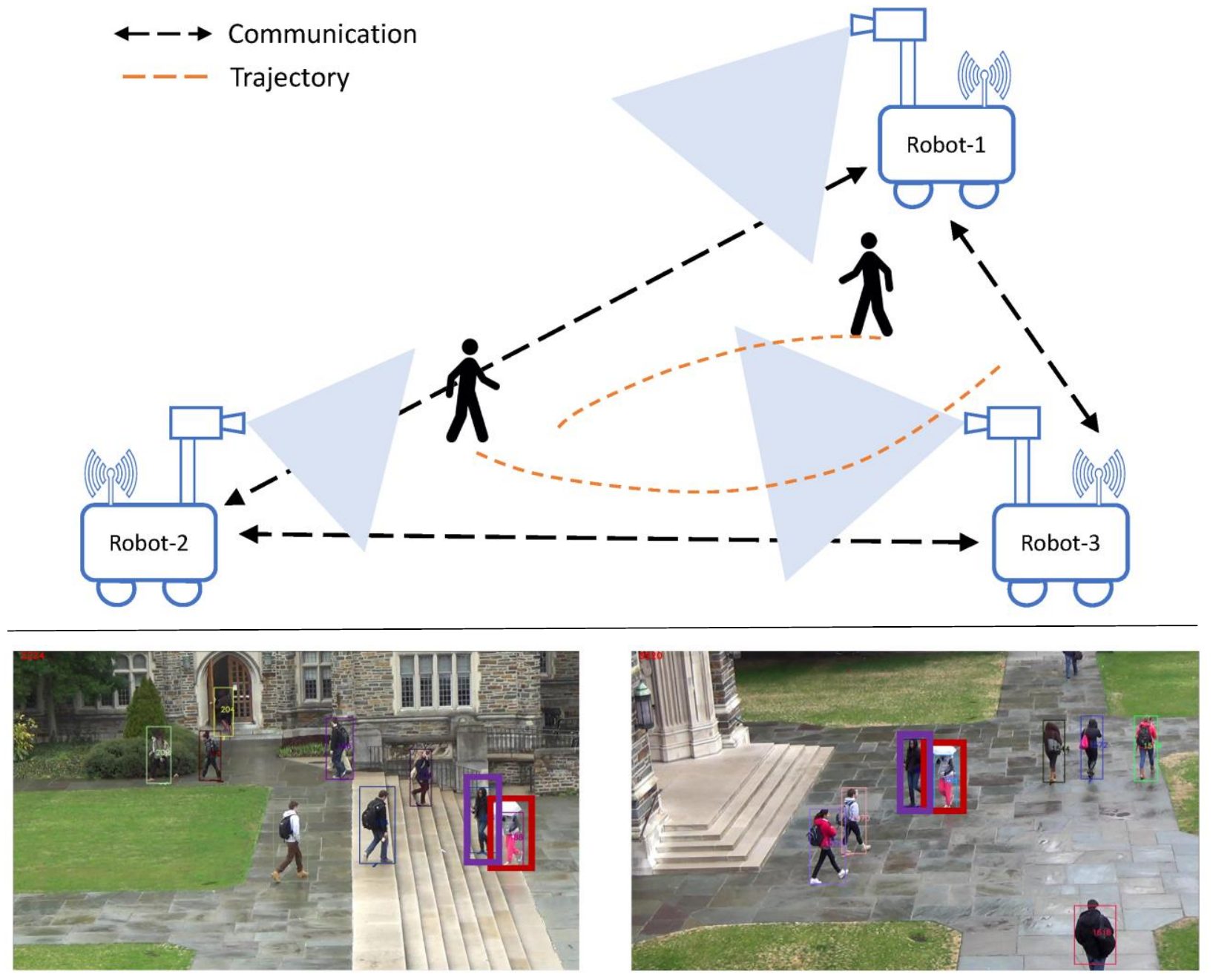}
    \caption{Illustration of multi-sensor setting with intermittent communication. The bottom figure shows our multi-sensor tracking results for a pair of images from camera 2 (left) and 5 (right) in DukeMTMC dataset, where some people appear in both cameras at different times. Their identities are correctly recognized across sensors (see red and violet bounding boxes).}
    \label{fig:overview}
\end{figure}

This problem is exacerbated when decentralized tracking is required, such as using a team of drones or ground mobile robots equipped with sensors for environmental monitoring under constrained communication resources. In fact, due to restrictions on communication bandwidth between robots, only limited data should be communicated between robots. Therefore, traditional MTMST approaches of sharing whole image galleries between robots is not valid in the decentralized setting. This work focuses on the more challenging problem of decentralized MTMST (D-MTMST), and proposes a communication-efficient tracking approach for multiple people tracking with networked sensors (Figure \ref{fig:overview}).

MTMST has a long history of development and initially relied on spatio-temporal information of tracked objects. Filtering approaches, such as Kalman filters and Particle filters, have been widely used for motion model-based tracking \cite{mittal2003m,breitenstein2011online}. With recent advances in computer vision, tracking-by-detection has become a popular paradigm for MTMST \cite{ess2009robust,ristani2018features},
which involves detecting objects of interest in each frame and linking object hypotheses into trajectories via data association algorithms. Such paradigm consists of two components: within-sensor people tracking and cross-sensor identity association. While the within-sensor tracking focuses on estimating pedestrians
trajectories within each sensor's FOV, cross-sensor identity association emphasizes on maintaining identity consistency for people moving across multiple sensors. 

Recent works on MTMST relies primarily on the combination of person re-IDentification (re-ID) and motion pattern information to achieve state-of-the-art tracking performance \cite{kuo2011does,tang2017multiple,ristani2018features}. With the use of deep neural networks, person re-ID algorithms have achieved impressive accuracy at visually identifying and associating people within and across different sensor frames. Motion pattern information, such as individual motion models and pedestrian flow patterns, places spatio-temporal constraints on data association to remove implausible tracks. Various approaches have been developed to build human motion patterns by incorporating prior information or online detection, including dynamic models \cite{kuo2011does}, probabilistic graphical models \cite{yang2014multi}, and neural networks \cite{narayan2018re}. 

While research on MTMST has achieved noticeable progress in recent years, most research focuses on a centralized tracking strategy, where each sensor's detections can be shared with all other sensors without communication constraints. There is however, little work in D-MTMST, addressing key challenges such as communication bandwidth, link robustness, cross-sensor data association, and scalability.

\begin{figure}
    \centering
    \includegraphics[width=.5\textwidth]{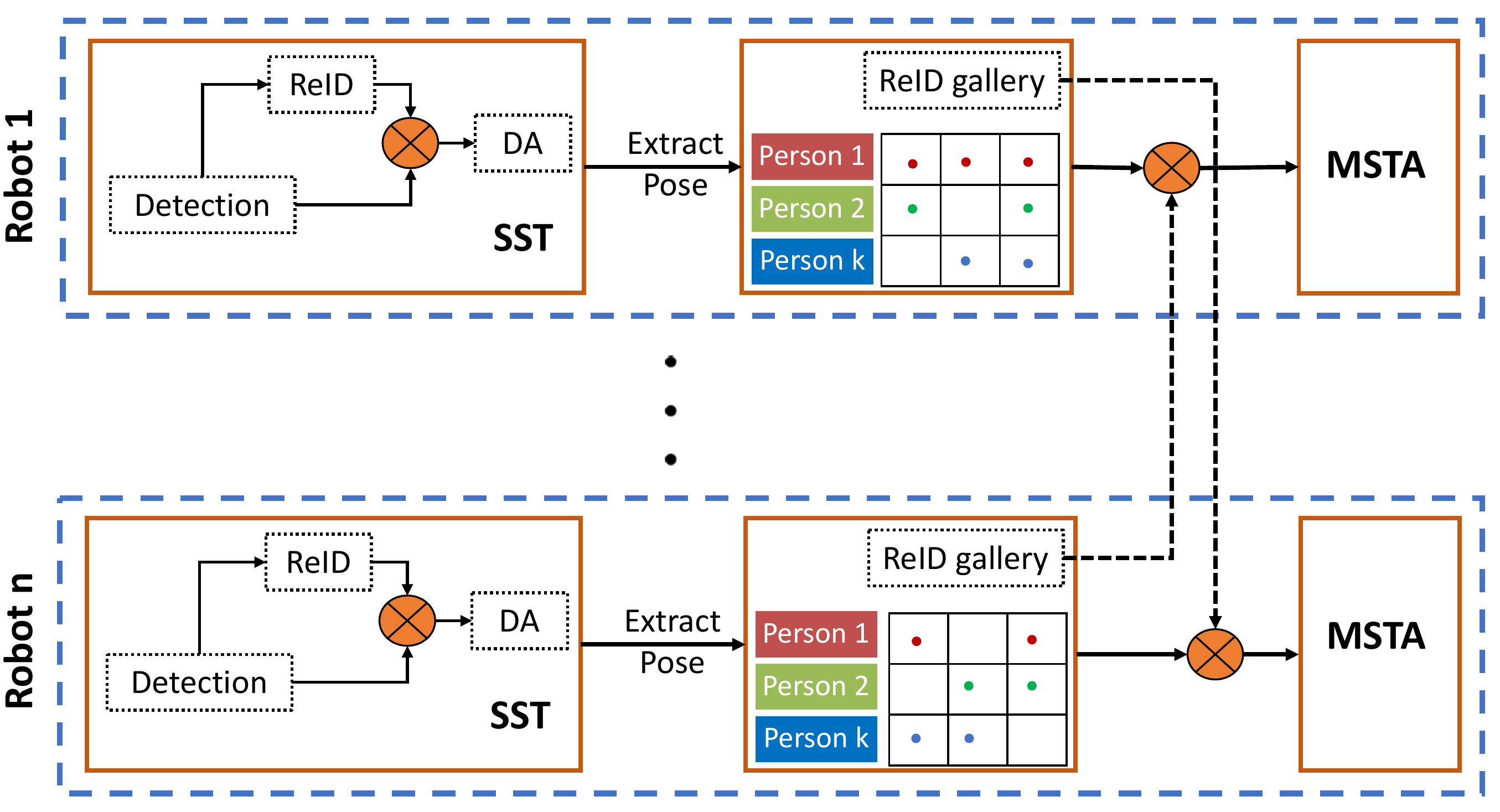}
    \caption{The proposed pipeline for D-MTMST in one batch processing cycle. Each sensor locally tracks people via the single-sensor tracking (SST) module. Person pose is extracted and utilized for generating an orientation-discriminative re-ID feature gallery, which is then shared among cameras. The shared gallery is used for multi-sensor track association (MSTA). After MSTA, a new tracking cycle started.}
    \label{fig:pipeline}
\end{figure}

This work proposes a communication-efficient approach for D-MTMST, as shown in Figure \ref{fig:pipeline}, based on our hypothesis that orientation-aware information will improve ReID and data association. This is accomplished by the following technical contributions: First, a person orientation descriptor is developed to characterize orientation from images. Second, an orientation-discriminative, appearance-based feature representation is proposed for track association, which significantly reduces communication overhead. 
Third, the orientation feature gallery is used to create cross-sensor track associations. The proposed D-MTMST approach is evaluated on the DukeMTMC dataset \cite{ristani2016performance}.

The paper is organized as follows. Section \ref{sec:related_work} reviews the related work. Section \ref{sec:prob_formulation} formulates the problem of D-MTMST. Section \ref{sec:orient_feature} subsequently describes the orientation-discriminative person feature representation for re-IDentification and identity association. Multi-sensor tracking approach is then presented in Section \ref{sec:multi_sensor_tracking}. Experiment results is shown and analyzed in Section \ref{sec:evaluation}. Finally, Section \ref{sec:conclusion} concludes the paper.

\section{Related Work}
\label{sec:related_work}

\textbf{Person re-ID.} Starting from manually designed features, such as color and texture descriptors \cite{farenzena2010person}, to features from metric learning \cite{koestinger2012large,liao2015person}, the state of the art in person re-IDentification counts on deep learning techniques, where person images are transformed by deep neural networks into high dimensional features in a way that distances between features of the same person are smaller than those between features of different people. Various techniques, such as data augmentation \cite{zhong2017random}, using body region \cite{zhao2017spindle}, and multi-scale feature learning \cite{qian2017multi}, have been proposed to improve the classification accuracy of person re-IDentification.

\textbf{Data Association.} Data association plays a central role in MTMST, which focuses on correctly associating detection to corresponding tracks. Data association is typically performed uses Bayes filtering approaches or multiple hypotheses techniques to sequentially link the detection with existing trajectories. While Multiple Hypothesis Tracking \cite{chen2017enhancing} typically maintains a few highly possible hypotheses according to deterministic branching decisions, filtering approaches keep track of a probability distribution via sampling \cite{miller2011efficient}.

\textbf{Multi-sensor Tracking.} While the goal of single-sensor target tracking is to associate consecutive detections with motion trajectories over time, multi-sensor tracking focuses on building consistent correspondences of trajectories from different sensors. Various approaches have been developed that utilize spatio-temporal relations for multi-sensor tracking in a decentralized framework \cite{ong2008decentralised,campbell2016distributed}. Combining with advances in computer vision, recent works \cite{ristani2018features,tesfaye2017multi} focus using both appearance cues and spatio-temporal relations. However, most works so far have assumed a centralized tracking framework, and little work is done for decentralized tracking using visual information.


\section{Problem Formulation}
\label{sec:prob_formulation}
Consider a system of $M$ camera sensors that are installed either on fixed bases or mobile robots. Assume that there are $N$ people in the environment. Each person $p\in\{1,\dots,N\}$ generates a sequence of trajectory in the $2$-D ground plane. Define $\mathbf{x}_p(t)\in\varmathbb{R}^{d_x}$ as the state of $p$th person along the trajectory at time instance $t$, where $d_x$ is the state dimension.
The sensor obeys a measurement model,
\begin{equation}
    \mathbf{z}_p(t)=\mathbf{g}(\mathbf{x}_p(t),\mathbf{w}),
\end{equation}
where $\mathbf{g}:\varmathbb{R}^{d_x}\times \varmathbb{R}^{d_x}\rightarrow\varmathbb{R}^{d_z}$ is the measurement model with the output dimension of $d_z$ and $\mathbf{w}$ is the measurement noise. The detection output $\mathbf{z}_p(t)=[\mathbf{z}^\text{app}_p(t),\mathbf{z}^\text{pos}_p(t)]\in\varmathbb{R}^{d_z}$ consists of the detected person image $\mathbf{z}^\text{app}_p(t)\in\varmathbb{R}^{d_a}$ and position $\mathbf{z}^\text{pos}_p(t)\in\varmathbb{R}^{d_p}$, with $d_a$ and $d_p$ being corresponding dimensions. Define $\mathcal{Z}^c(t)=\{\mathbf{z}^c_j(t)|j=1,\dots,N^c(t)\}$ as the set of all $N^c(t)$ object detections observed by the sensor $c\in\{1,\dots,M\}$ at the time instance $t$. 
Note that the number of detections is time varying as objects constantly enter and leave a sensor's field of view (FOV).



\subsection{Probabilistic Tracking}
A probabilistic tracking framework is used
to estimate the joint probability distribution over discrete data assignments and continuous person states. 
Define $\mathcal{X}=\{\mathbf{x}_p(t)|p=1,\dots,N,\,t=1,\dots,T\}$ as the set of all pedestrians' states at time $1$ through $T$ and 
$\mathcal{Z}=\cup_{c=1,\dots,M,\,t=1,\dots,T}\mathcal{Z}^c(t)$ 
as the set of all sensors' measurements. 
Notation $T$ represents the length of a fusion batch, where multi-sensor data association is conducted after running single-sensor tracking for $T$ time steps (as shown in Figure \ref{fig:pipeline}). 
Let $\mathbf{\Theta}^c_{\text{ws}}=\{\theta_{pj}(t)|p=1,\dots,N, \,j=1,\dots,N^c(t),\, t=1,\dots,T\}$
represent the data association in $c$th sensor, where $\theta_{pj}(t)\in \{0,1\}$ is an indicator variable connecting each measurement to its object of origin. The $\theta_{pj}(t)=1$ if and only if $\mathbf{z}_j(t)$ is the measurement of $p$th person.
Also let $\mathbf{\Theta}_{\text{cs}}=\{\bar{\theta}_{j_{c_1}j_{c_2}}|j_{c_1}=1,\dots,N^{c_1},\,j_{c_2}=1,\dots,N^{c_2}\}$ represent the pairwise trajectory association, where $\bar{\theta}_{j_{c_1}j_{c_2}}\in\{0,1\}$ equals $1$ if and only if the $j_{c_1}$th trajectory in sensor $c_1$ and $j_{c_2}$th trajectory in sensor $c_2$ correspond to the same person. The $N^{c_1}$ and $N^{c_2}$ represent the total number of trajectories in each sensor that is to be estimated.


The D-MTMST computes the maximum-a-posteriori (MAP) of the jointly probability distribution of the trajectories and data association , i.e. 
\begin{equation}
    \mathbf{\Theta}^*_{\text{cs}},\mathbf{\Theta}^*_{\text{ws}},\mathcal{X}^*=\arg\max_{\mathbf{\Theta}_{\text{cs}},\mathbf{\Theta}_{\text{ws}},\mathcal{X}}P(\mathbf{\Theta}_{\text{cs}},\mathbf{\Theta}_{\text{ws}},\mathcal{X}|\mathcal{Z}).
\end{equation}
The joint probability distribution can be factorized as
\begin{equation}
    \begin{split}
    \label{eqn:prob_tracking}
        P(\mathbf{\Theta}_{\text{cs}},\mathbf{\Theta}_{\text{ws}},\mathcal{X}|\mathcal{Z})&=P(\mathbf{\Theta}_{\text{cs}}|\mathbf{\Theta}_{\text{ws}},\mathcal{X},\mathcal{Z})P(\mathbf{\Theta}_{\text{ws}},\mathcal{X}|\mathcal{Z})\\
        &=P(\mathbf{\Theta}_{\text{cs}}|\mathbf{\Theta}_{\text{ws}},\mathcal{X},\mathcal{Z})P(\mathbf{\Theta}_{\text{ws}}|\mathcal{Z})P(\mathcal{X}|\mathbf{\Theta}_{\text{ws}},\mathcal{Z}).
    \end{split}
\end{equation}

The term $P(\mathcal{X}|\mathbf{\Theta}_{\text{ws}},\mathcal{Z})$ represents the conditional probability of person states given data associations $\mathbf{\Theta}_{\text{ws}}$ and measurements $\mathcal{Z}$, which can be computed using filtering algorithms such as the Kalman filter \cite{thrun2005probabilistic}. The cross-camera association term $P(\mathbf{\Theta}_{\text{cs}}|\mathbf{\Theta}_{\text{ws}},\mathcal{X},\mathcal{Z})$ is a discrete distribution over association pairs.

The term $P(\mathbf{\Theta}_{\text{ws}}|\mathcal{Z})$ is a probability distribution over data association given measurements. Since in practice this term involves a large number of association permutations to be estimated, the Particle filter \cite{thrun2005probabilistic} is used for approximating $P(\mathbf{\Theta}_{\text{ws}}|\mathcal{Z})$. The particles are drawn according to a proposal density $q(\mathbf{\Theta}_{\text{ws}}|\mathcal{Z})$ selected for efficient sampling. The true density $P(\mathbf{\Theta}_{\text{ws}}|\mathcal{Z})$ is then approximated by the weighted sum of samples,
\begin{equation}
    P(\mathbf{\Theta}_{\text{ws}}|\mathcal{Z})\approx \sum_i^{n_p} w_i\delta(\mathbf{\Theta}_{\text{ws}}-\mathbf{\Theta}_{\text{ws},i}),
\end{equation}
where $n_p$ is the number of particles and $w_i$ is the weight of the $i$th particle $\mathbf{\Theta}_{\text{ws},i}$, with $w_i$ defined as
\begin{equation*}
    w_i=\frac{P(\mathbf{\Theta}_{\text{ws},i}|\mathcal{Z})}{q(\mathbf{\Theta}_{\text{ws},i}|\mathcal{Z})}.
\end{equation*}
The $\delta(\cdot)$ denotes the Kronecker function.
The conditional probability density (\ref{eqn:prob_tracking}) is consequently approximated as
\begin{equation}
    P(\mathbf{\Theta}_{\text{ws}},\mathcal{X}|\mathcal{Z})\approx \sum\limits_i^{n_p}  w_i\delta(\mathbf{\Theta}_{\text{ws}}-\mathbf{\Theta}_{\text{ws},i})P(\mathcal{X}|\mathbf{\Theta}_{\text{ws},i},\mathcal{Z}).
\end{equation}
and each particle stores one possible data assignment history. This use of combining Monte Carlo sampling methods with closed-form parametric filters to estimate a factorized probability density is known as the Rao-Blackwellized Particle filter \cite{miller2011efficient}. 

\section{Orientation-Discriminative Person Feature Representation}
\label{sec:orient_feature}
This section proposes an orientation-discriminative person re-ID feature representation that leads to improved re-ID performance and efficient data sharing. Pedestrian feature extraction by using person re-ID and pose estimation approaches is first described in Section \ref{subsec:re-ID-pose-est}. Then the proposed orientation estimation and clustering is presented in Section \ref{subsec:orient_est}. Orientation-discriminative feature representation is subsequently introduced in Section \ref{subsec:orient_feat_rep}.


\subsection{Re-Identification and Pose Estimation} 
\label{subsec:re-ID-pose-est}
re-IDentification (re-ID) is defined as the task of establishing correspondence between images of a person taken from different cameras, where a stable identity is assigned to different instances of the same person. In the traditional re-ID setting, a \textit{gallery} of person images and a set of \textit{query} frames are provided, and the goal of person re-ID algorithms is to retrieve gallery images that contain the same person as a query frame. To achieve this, a person re-ID algorithm generates a feature vector, known as the \textit{embedding}, for each person image in the way that features belonging to the same person are closer than that belonging to different people. This work uses a state-of-the-art person re-ID algorithm, namely Deep Any-time re-ID (DaRe) \cite{wang2018resource}, for re-ID feature extraction. Specifically, define the neural network of DaRe as a function $\mathbf{f}:\varmathbb{R}^{d_a}\rightarrow \varmathbb{R}^{n_f}$, where $n_f$ is the feature dimension. Then re-ID feature $\boldsymbol{\eta}\in\varmathbb{R}^{n_f}$ is defined as
\begin{equation*}
    \boldsymbol{\eta}_p=\mathbf{f}(\mathbf{z}^\text{app}_p),\,\forall p =1,\dots,N.
\end{equation*}

This work leverages pose information in addition to person re-ID features. Specifically, OpenPose \cite{cao2018openpose} is used for pose estimation, which generates a non-parametric representation of keypoint association that encodes the position and orientation of human limbs to jointly learn human part locations and their associations. Our work feeds the $2$D bounding box detections from an object detector to OpenPose to retrieve $2$D keypoint localization, along with the confidence values associated with each keypoint. The keypoints are used to estimate person orientations, which are then utilized for grouping  re-ID features into discrete orientation bins, as described in the following subsections.

\subsection{Orientation Descriptor} 
\label{subsec:orient_est}
Characterizing orientation from $2$D keypoint locations is a challenging task due to the lack of depth information. To achieve this, this paper proposes a metric $r_{S2T}$, called \textit{S2T ratio}, to quantify the ratio of the shoulder width $p_w$ to the torso length $p_h$ in the image frame, defined as follows,
\begin{gather}
\label{eqns:S2T_def}
    r_{S2T}(\mathbf{z}^\text{app}_p) = \frac{p_w(\mathbf{z}^\text{app}_p)}{p_h(\mathbf{z}^\text{app}_p)}\in\varmathbb{R},\\
    p_w(\mathbf{z}^\text{app}_p) = \frac{(c_{Rs} + c_{Ls}) (x_{Rs} - x_{Ls}) + (c_{Rh} + c_{Lh}) (x_{Rh} - x_{Lh})}{c_{Rs} + c_{Ls} + c_{Rh} + c_{Lh}},\nonumber\\ 
    p_h(\mathbf{z}^\text{app}_p) = \frac{(c_{Rs} + c_{Rh}) (y_{Rs} - y_{Rh}) + (c_{Ls} + c_{Lh}) (y_{Ls} - x_{Lh})}{c_{Rs} + c_{Ls} + c_{Rh} + c_{Lh}},\nonumber
\end{gather}
where $(x_{Rs}, y_{Rs})$, $(x_{Ls}, y_{Ls})$ are right and left shoulders' coordinates in image $\mathbf{z}^\text{app}_p$ respectively, and $(x_{Rh}, y_{Rh})$, $(x_{Lh}, y_{Lh})$ are right and left hips' coordinates respectively. The notation `$c$' denotes the confidence level associated with corresponding keypoints, given by OpenPose.

\begin{figure}
    \centering
    \includegraphics[width=.5\textwidth]{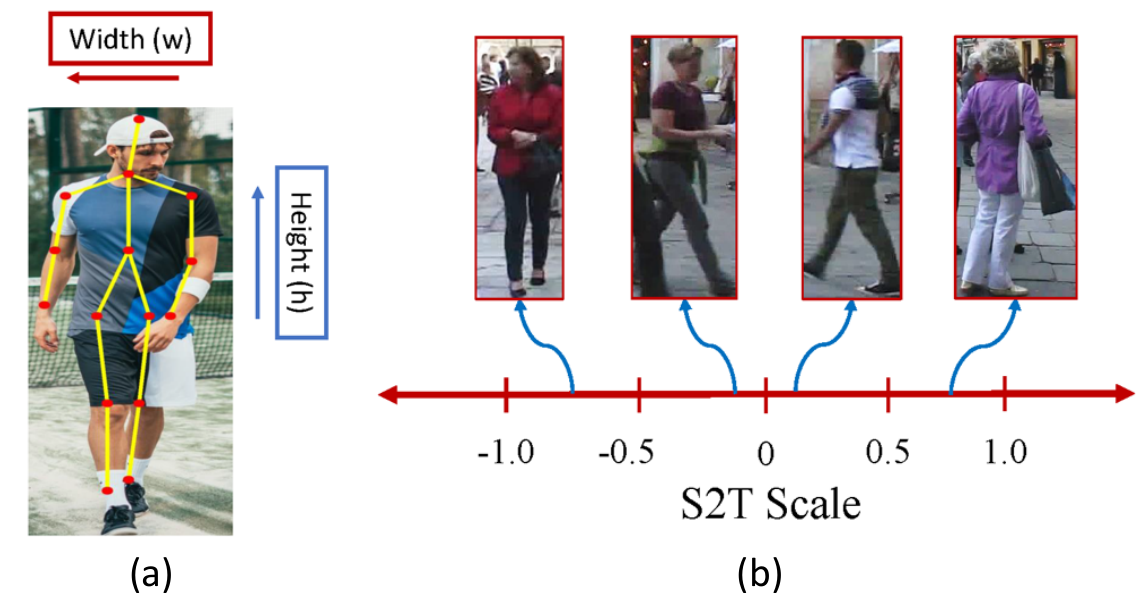}
    \caption{(a) Illustration of the width and height used for calculating S2T ratio using detected keypoints. (b) Images belonging to different regions in the S2T scale. \textit{Best viewed in color.}}
    \label{fig:S2T}
\end{figure}

The S2T ratio is a \textbf{signed} ratio of body width to body height in the image frame and provides a tool to characterize person orientation with respect to the vertical axis, as shown in Figure~\ref{fig:S2T}(a). Having a high (positive) value of $r_{S2T}$ means that the person is facing away from the camera, while having a low (negative) value of $r_{S2T}$ means that the person is facing towards the camera, as illustrated in Figure \ref{fig:S2T}(b). Since it is based upon relative positions of keypoints, this metric is robust towards bounding box misalignment of the detector. However, the greatest motivation to use S2T ratio is because it is based upon the major keypoints of the torso, which are accessible in most bounding box detections provided by the detector. In addition, weighing the coordinates with their confidence values makes $r_{S2T}$ robust to small level occlusion. 

\textbf{Remark}.
The full-vector of detected keypoints of a person are sparsely distributed in the high-dimensional space. Ideally, one could cluster them in the high dimensional space to preserve more general pose variations, rather than just orientation. However, since the detections are not perfect, some keypoints are missing in practice, which adversely affects the clustering approach. The keypoints used in S2T ratio are usually present across most detections, which contributes to its reliability and robustness. 

\subsection{Orientation-discriminative Feature Representation} 
\label{subsec:orient_feat_rep}
In person re-ID, the common strategy for associating person detections is retaining features of all video frames for each person to keep maximal information for identification. This strategy, however, cannot work for decentralized multi-sensor video tracking, since sharing all features between sensors can cause high communication overhead, which does not scale with tracks, people or sensors. In fact, assuming that $N$ is the people number and $F$ denotes the total number of frames of a sensor, the sensor's gallery is $O(NF)$ when all features are retained, which renders this strategy unscalable for a long duration.

To address this issue, a novel feature representation is proposed based on person orientation, which achieves a space complexity of $O(N)$. Define a sequence of $L+1$ real scalars $b_0,\dots,b_L\in\varmathbb{R}$ with $b_0<\dots<b_L$ and $L$ bins where the $l$th bin corresponds to the interval $B_l=[b_{l-1},b_l),\,l=1,\dots,L$. Consider the set of person $p$'s images of which the S2T ratio belongs to $B_l$, i.e. $\mathbf{I}_{p,l} = \{\mathbf{z}^\text{app}_p|r_{S2T}(\mathbf{z}^\text{app}_p)\in B_l\}$. Define the average feature of all images in $\mathbf{I}_{p,l}$ as the representation feature for the $l$th bin, i.e.
\begin{equation}
\label{eqn:bin_feature}
    \boldsymbol{\eta}_{B_{p,l}}=\frac{1}{|\mathbf{I}_{p,l}|}\sum\limits_{\mathbf{z}\in \mathbf{I}_{p,l}} \mathbf{f}(\mathbf{z}),
\end{equation}
where $|\cdot|$ denotes the set cardinality. The vector $\boldsymbol{\eta}_{B_{p}}=[\boldsymbol{\eta}_{B_{p,1}}^T,\dots,\boldsymbol{\eta}_{B_{p,L}}^T]^T$, obtained via this binning process, composes the orientation-discriminative feature representation of person $p$.

This feature representation achieves a desirable balance between representativeness and compactness, and is therefore particularly suitable for multi-sensor scenarios. On one hand, person appearance significantly changes due to the variation of sensor view angles. The orientation-discriminative feature representation maintains running averages within multiple orientation bins for each person and therefore preserves the appearance information associated with different orientations of a person. On the other hand, such feature representation reduces the gallery size to $O(N)$, which is desirable not only for quick correspondence computations, but eases data transmission between devices as well. In this case, the communication data between sensors will be $O(N)$, linear in the number of active tracks and independent of recording time.

\section{Data Association for Multi-Sensor Tracking}
\label{sec:multi_sensor_tracking}
\subsection{Probabilistic Within-sensor Tracking}
The task of within-camera data association is to assign new detections to existing tracks or create new tracks if the detected people are determined to be new identities.
To compute the spatial likelihood $P(\mathcal{X}|\mathbf{\Theta},\mathcal{Z})$ in (\ref{eqn:prob_tracking}), this work assumes a linear motion model of people in the image plane and uses the Kalman filter for estimation. Specifically, the person state consists of six dimensions, i.e. $\mathbf{x}_p=[x_p,\dot{x}_p,y_p,\dot{y}_p,\tau_{w,p},\tau_{h,p}]\in\varmathbb{R}^6$, where $x_p,\dot{x}_p,y_p,\dot{y}_p$ denote the position and velocity in the $2$D image plane and $\tau_w,\tau_h$ represent the bounding box size of each detected person. The motion model and sensor measurement model are defined as
\begin{subequations}
    \begin{align}
        \mathbf{x}_p(t)&=A\mathbf{x}_p(t-1)+\mathbf{v},\\
        \mathbf{z}^\text{pos}_p(t-1)&=C\mathbf{x}_p(t-1)+\mathbf{w},\\
         A=\begin{bmatrix}
        0 & \delta t & 0 & 0 & 0 & 0\\
        0 & 1 & 0 & 0 & 0 & 0\\
        0 & 0 & 0 & \delta t & 0 & 0\\
        0 & 0 & 0 & 1 &0  & 0\\
        0 & 0 & 0 & 0 & 0 & 0\\
        0 & 0 & 0 & 0 & 0 & 0\\
        \end{bmatrix},\quad
        & C = \begin{bmatrix}
        1 & 0 & 0 & 0 & 0 & 0\\
        0 & 0 & 1 & 0 & 0 & 0\\
        0 & 0 & 0 & 0 & 1 & 0\\
        0 & 0 & 0 & 0 & 0 & 1\\
        \end{bmatrix},\nonumber
    \end{align}
\end{subequations}
where $\delta t$ is the discretization time interval.
The $v\in\varmathbb{R}^6$ and $w\in\varmathbb{R}^4$ denote the process and measurement Gaussian noise, respectively. 

The likelihood term $P(\mathbf{\Theta}|\mathcal{Z})$ plays a vital role in data association. Both person position $\mathbf{z}^\text{pos}_j$ and appearance $\mathbf{z}^\text{app}_j$ are utilized to compute this conditional probability. For clarity, consider the conditional probability $P(\theta_{pj}(t)|\mathbf{z}_{j}(t)))$. Assuming a conditional independence between position and appearance given the identity of the person, $P(\theta_{pj}(t)|\mathbf{z}_{j}(t)))$ can be decomposed as follows,
\begin{subequations}
\begin{align}
P(\theta_{pj}(t)|\mathbf{z}_{j}(t))) & = P(\theta_{pj}(t)|\mathbf{z}^\text{pos}_{j}(t),\mathbf{z}^\text{app}_{j}(t))\\
& = \alpha P(\mathbf{z}^\text{pos}_{j}(t),\mathbf{z}^\text{app}_{j}(t)|\theta_{pj}(t))P(\mathbf{z}^\text{pos}_{j}(t),\mathbf{z}^\text{app}_{j}(t))\label{subeqn:likeli_bayes}\\
& = \alpha P(\mathbf{z}^\text{pos}_{j}(t)|\theta_{pj}(t))\cdot P(\mathbf{z}^\text{app}_{j}(t)|\theta_{pj}(t))\cdot\nonumber\\
~&\quad P(\mathbf{z}^\text{pos}_{j}(t),\mathbf{z}^\text{app}_{j}(t))\label{subeqn:likeli_cond_indep}
\end{align}
\end{subequations}
where (\ref{subeqn:likeli_bayes}) is from the Bayes rule and $\alpha$ is the normalizing constant. Equation (\ref{subeqn:likeli_cond_indep}) is based on the conditional independence assumption. The spatial likelihood $P(\mathbf{z}^\text{pos}_{j}(t)|\theta_{pj}(t))$ makes use of human motion model and the appearance likelihood $P(\mathbf{z}^\text{app}_{j}(t)|\theta_{pj}(t))$ accounts for the visual similarity of detections.

Given the predicted state $\mathbf{x}_p(t)$ (via Kalman filter) and the detected position $\mathbf{z}^\text{pos}_j(t)$, the spatial likelihood is computed by using the softmin function over the Mahalanobis distance between them,

\begin{equation}
P((\mathbf{z}^\text{pos}_j(t)|\theta_{pj}(t)) = \frac{\exp(-\beta\|\mathbf{z}^\text{pos}_j(t) - C\mathbf{x}_p(t)\|_{M})}{\sum\limits_{p=1,\dots,N}\exp(-\beta\|\mathbf{z}^\text{pos}_j(t) - C\mathbf{x}_p(t)\|_{M})},
\label{eqn: pos_like}
\end{equation}
where $\beta$ is a tuning parameter. 

The appearance likelihood is computed by comparing detection $\mathbf{z}^\text{app}_j(k)$ and the features of person $p$ in the feature space. Specifically, the $L_2$ norm between $\mathbf{f}(\mathbf{z}^\text{app}_j(k))$ and feature $\boldsymbol{\eta}_{B_{p,l}}$ is computed and transformed via a softmin function,
\begin{equation}
\begin{split}
P((\mathbf{z}^\text{app}_j(t)|\theta_{pj}(t)) = \frac{\exp(-\gamma\min\limits_{l=1,\dots,T}\|\mathbf{f}(\mathbf{z}^\text{app}_j(k))-\boldsymbol{\eta}_{B_{p,l}}\|_2)}{\sum\limits_{p=1,\dots,N}\exp(-\gamma\min\limits_{l=1,\dots,T} \|\mathbf{f}(\mathbf{z}^\text{app}_j(k))-\boldsymbol{\eta}_{B_{p,l}}\|_2)}
\end{split}
\end{equation}
where $\gamma$ is a tuning parameter.

Though here the softmax function is used to transform a matching score into range $[0,1]$, other techniques could also be used to transform into probability, such as by modeling the genuine and impostor PDFs \cite{eisenbach2015user}.

\subsection{Cross-sensor Data Association}
Computing MAP of $P(\mathbf{\Theta}_{\text{cs}},\mathbf{\Theta}_{\text{ws}},\mathcal{X}|\mathcal{Z})$, namely $\{\mathbf{\Theta}^*_{\text{cs}},\mathbf{\Theta}^*_{\text{ws}},\mathcal{X}^*\}$, is a challenging task due to the combinatorial number of possible associations of sensors' trajectories. In this work, we take an approximate approach to compute MAP in a sequential manner. 
To be specific, $c$th sensor computes the locally most likely particle (data association and trajectories), $\boldsymbol{\Theta}^{c*}_{ws}$, to other sensors, i.e. 
\begin{equation}
    \boldsymbol{\Theta}^{c*}_{ws} = \arg\max\limits_{\boldsymbol{\Theta}^c_{ws}} P(\boldsymbol{\Theta}^c_{ws}|\mathcal{Z}^c),\,\forall c = 1,\dots,M,
\end{equation}
where the superscript $c$ denotes the $c$th camera.
Once the association is determined, the trajectory estimation $\mathcal{X}^*$ can be computed from Kalman filter. The corresponding feature bins $\boldsymbol{\eta}^{c*}_{B_l},l=1,\dots,T$ can be computed from associated detections.

The $\mathbf{\Theta}^*_{\text{cs}}$ is computed as follows,
\begin{equation}
\label{eqn:map_cs}
    \mathbf{\Theta}^*_{\text{cs}}=\arg\max_{\mathbf{\Theta}_{\text{cs}}} P(\mathbf{\Theta}_{\text{cs}}|\mathbf{\Theta}^*_{\text{ws}},\mathcal{X}^*,\mathcal{Z}).
\end{equation}
Recall that cross-sensor data association in our work only relies on re-ID feature gallery. Therefore, (\ref{eqn:map_cs}) can be reduced to 
\begin{equation}
\label{eqn:map_cs2}
    \mathbf{\Theta}^*_{\text{cs}}=\arg\max_{\mathbf{\Theta}_{\text{cs}}} P(\mathbf{\Theta}_{\text{cs}}|\mathbf{\Theta}^*_{\text{ws}})
\end{equation}
Equation (\ref{eqn:map_cs2}) is solved by sequentially associating person identities between  pairs of two sensors. Consider sensor $c_1$ and $c_2$ and let $\boldsymbol{\eta}^{c_1}_{B_{j_1}}$ and $\boldsymbol{\eta}^{c_2}_{B_{j_2}}$ represent the binned feature gallery of trajectory $j_1$ and $j_2$. The similarity score between these two trajectories is defined via softmax function as,
\begin{equation}
    A_{j_1j_2}=\exp{(-\gamma \|\boldsymbol{\eta}^{c_1}_{B_{j_1}}-\boldsymbol{\eta}^{c_2}_{B_{j_2}}\|_2)},
\end{equation}
where $\gamma$ is a tuning parameter.
The joint probability $P(\mathbf{\Theta}_{\text{cs}}|\mathbf{\Theta}^*_{\text{ws}})$ can be represented as
\begin{equation}
    \begin{split}
    \label{eqn:cs_prob}
        P(\mathbf{\Theta}_{\text{cs}}|\mathbf{\Theta}^*_{\text{ws}})&=\prod_{\substack{j_{c_1}=1:N^{c_1}\\j_{c_2}=1:N^{c_2}}}A^{\bar{\theta}_{j_{c_1}j_{c_2}}}_{j_{c_1}j_{c_2}}\\
        &=\exp{\left(-\gamma \sum_{\substack{j_{c_1}=1:N^{c_1}\\j_{c_2}=1:N^{c_2}}}\bar{\theta}_{j_{c_1}j_{c_2}}\|\boldsymbol{\eta}^{c_1}_{B_{j_{c_1}}}-\boldsymbol{\eta}^{c_2}_{B_{j_{c_2}}}\|_2\right)}
    \end{split}
\end{equation}
where $\bar{\theta}_{j_{c_1}j_{c_2}}$ denotes the association variable defined in Section \ref{sec:prob_formulation}.
Note the optimization in (\ref{eqn:map_cs2}) is under the constraint that a trajectory in one sensor can be associated with at most one trajectory in another sensor. To incorporate this constraint, maximizer of (\ref{eqn:cs_prob}) is computed using the Hungarian algorithm \cite{kuhn1956variants} to solve a maximum weight matching problem in bipartite graphs. Multi-sensor data association is conducted in a sequential way, between two sensors at each time step. After cross-sensor association, sensors' binned feature representation is merged with those from other sensors, which will used for multi-sensor tracking in the next batch processing cycle.

\textbf{Remark.} Our work uses the image plane position as states of the pedestrians. However, if the camera parameters and states are known, the world frame coordinates can be easily generated by a simple projection operation from image to ground plane. Also, the presented cross-camera data association strategy mainly relies on person re-ID feature. This is critical for applications where sensors have non-overlapping FOVs and sensors' ego-states are not available. For situations where sensors' FOVs overlap, position information can be utilized for data association across cameras. In this case, the presented approach is still valuable and can be further assisted by the position information of tracked people across cameras. 

\section{Evaluation}
\label{sec:evaluation}
The proposed D-MTMST approach is evaluated in both person re-ID and tracking performance. Section \ref{subsec:re-ID} evaluates the orientation-discriminative feature representation for person re-ID. Then probabilistic tracking using a single camera is tested in Section \ref{subsec:single-sensor}. Multi-sensor decentralized tracking is evaluated in Section \ref{subsec:multi-sensor}.

\subsection{re-ID}
\label{subsec:re-ID}
The orientation-discriminative feature gallery is evaluated on three pedestrian tracking datasets, PETS09-S2L1, ADL-Rundle-8, and Venice-2 \cite{MOTChallenge2015}, for person re-ID. These datasets include a wide variety of conditions that are encountered in robotics applications, including occlusions (Venice-2), low illumination (ADL-Rundle-8) and sensor motion (ADL-Rundle-8).
Importantly, pedestrian tracking videos differ from person re-ID datasets in the large number of detections per person; this emphasizes the need for a compact representation of the features for computational efficiency. 

\begin{figure}
    \centering
    \includegraphics[trim= 250 120 250 110, clip,width=.5\textwidth]{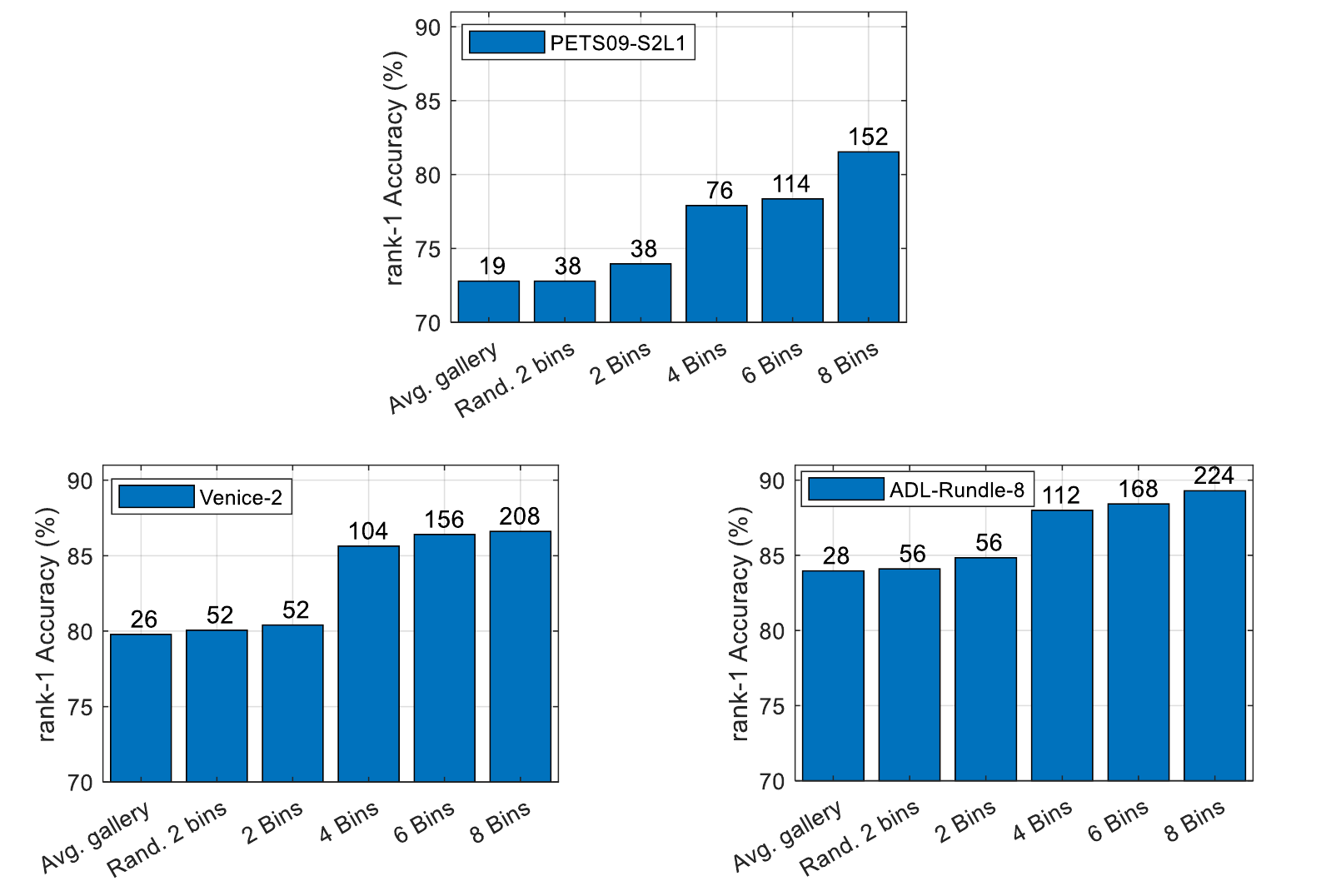}
    \caption{re-ID performance comparison for different feature representations, evaluated on different datasets. The total number of features stored in the gallery is shown at the top of the bar.}
    \label{fig:reid_bin_superior}
\end{figure}

To perform re-ID on tracking datasets, the videos are pre-processed to build the gallery and the query set. The ground-truth detections are utilized to extract people bounding boxes from videos, which are then randomly divided into gallery (80\%) and query (20\%) sets. The average rank-1 accuracy is used as the evaluation metric, defined as the proportion of query images which share the same ID with their corresponding nearest neighbor in the gallery. The DaRe network is trained on Market-1501 dataset \cite{zheng2015scalable} for re-ID, and is used for converting images into feature embeddings, without any fine-tuning on our test dataset. The test results are presented in Figure \ref{fig:reid_bin_superior}.

In the average gallery approach, only the average of all the features corresponding to a person is stored in the gallery; thus, leading to a much smaller gallery size. In the random two-bin gallery approach, all the features are randomly divided into two bins and the average feature within each bin is stored. In the orientation-based binning approach, first, OpenPose is used for extracting keypoints from detection bounding boxes and $S2T$ ratio is estimated. This is followed by clustering the features into $L$ bins by performing k-means clustering, based upon the $S2T$ ratio. Finally, the average feature in each bin is kept as a representative of that bin.

As shown in Figure \ref{fig:reid_bin_superior}, the random-binning approach leads to negligible improvement in re-ID performance over the average-gallery approach. While, with just two orientation-based bins, we achieve a better rank-1 accuracy than the average-gallery and random-bin approach. This supports our hypothesis that dividing the gallery into bins, based upon $S2T$ ratio of the person, helps in preserving semantic information such as orientation, which would otherwise be lost in the average gallery operation. To further study the scalability of our approach, we show the rank-1 accuracy with different numbers of bins in Figure \ref{fig:reid_bin_superior}. We observe a consistent improvement in the accuracy with increasing number of bins, across all the datasets. By adding just 8 bins, we get an average boost of \textbf{7\%} in rank-1 accuracy. This highlights an attractive aspect of our approach which allows us to trade-off between accuracy and bandwidth/computational requirements. 

It should be noted that, using the full gallery approach obviously results in better re-ID performance, achieving a rank-1 accuracy of 93.28\%, 97.84\% and 98.21\% for the respective datasets: PETS09-S2L1, Venice-2 and ADL-Rundle-8. However, this performance comes at a cost of significant memory overhead of having a large gallery size i.e. 3713, 5707 and 5418 features for the respective datasets. In the D-MTMST setting, there are two main drawbacks to this approach: first, it poses significantly higher bandwidth requirement for communication, and second, the larger gallery increases computational costs for data association which renders the method infeasible for prolonged tracking. 
 



\begin{table*}
\centering
\caption{Single-sensor Tracking Performance on Three Datasets}
\begin{tabular}{|l|l|l|l|l|l|l|l|l|l|}
\hline
~ & \textbf{Metrics} & \textbf{IDF1} & \textbf{IDP} & \textbf{IDR} & \textbf{FP} & \textbf{FN} & \textbf{IDs} & \textbf{MOTA} & \textbf{MOTP} \\ \hline
\multirow{4}{*}{\rotatebox[origin=c]{90}{PETS09}} & Only reID & 44.7 & 40.0 & 50.6 & 2330 & 1136 & 97 & 20.4 & 71.2 \\ \cline{2-10}
& Only reID with Ori. & 50.9 & 49.8 & 51.9 & 1798 & 1606 & 65 & 22.5 & 70.6\\ \cline{2-10}
& Pos. + Avg. reID feature & 82.7 & 79.1 & 86.7 & 905 & 475 & 8 & 69.0 & 72.2 \\ \cline{2-10}
& Pos. + reID with Ori. & 87.9 & 86.0 & 89.9 & 618 & 410 & 6 & 76.9 & 75.0 \\ \hline
\multirow{4}{*}{\rotatebox[origin=c]{90}{ADL-Run.}}& Only reID & 25.0 & 41.8 & 17.8 & 1396 & 5294 & 49 & 0.6 & 70.5\\ \cline{2-10}
& Only reID with Ori. & 30.7 & 48.2 & 22.5 & 1232 & 4847 & 28 & 10.0 & 70.0 \\ \cline{2-10}
& Pos. + Avg. reID feature & 52.4 & 48.3 & 57.2 & 3451 & 2193 & 35 & 16.3 & 70.5 \\ \cline{2-10}
& Pos. + reID with Ori. & 51.4 & 48.5 & 54.7 & 3111 & 2242 & 33 & 20.6 & 70.6\\ \hline
\multirow{4}{*}{\rotatebox[origin=c]{90}{Venice-2}} & Only reID & 26.1 & 40.3 & 19.3 & 1708 & 5435 & 51 & -0.7 & 71.1\\ \cline{2-10}
& Only reID with Ori. & 34.9 & 54.2 & 25.8 & 1125 & 4865 & 18 & 15.9 & 72.3 \\ \cline{2-10}
& Pos. + Avg. reID feature & 50.7 & 48.3 & 53.4 & 3168 & 2424 & 33 & 21.2 & 72.1 \\ \cline{2-10}
& Pos. + reID with Ori. & 50.4 & 48.7 & 52.3 & 3025 & 2498 & 30 & 22.2 & 72.6 \\ \hline
\end{tabular}
\label{tbl:SCT}
\end{table*}

\subsection{Single-sensor Tracking}
\label{subsec:single-sensor}
The proposed probabilistic tracking algorithm is evaluated on the same datasets. We have used the standard CLEAR MOT metrics \cite{bernardin2008evaluating} i.e. Multiple Object Tracking Accuracy (MOTA) and Multiple Object Tracking Precision (MOTP), for evaluating single-sensor tracking. MOTA combines three sources of error i.e. false positives (FP), false negatives (FN) and identity switches (IDs). MOTP quantifies the dissimilarity between the true positives and their corresponding ground truth targets. In addition, we also utilize IDF1, IDP and IDR  \cite{ristani2016performance} as a measure of identity matching performance. 

Four different approaches are compared. The first two are appearance-only approaches that use averaged re-ID features or orientation-discriminative features for tracking, without using position information associated with the detections. The other two methods use both, appearance and position, for tracking. In our implementation, the detections were derived from Mask R-CNN, which is pre-trained on COCO \cite{lin2014microsoft} dataset, with a threshold confidence level of 0.91. We have used $n_p=20$ particles for within-sensor data association and the orientation-binning approach uses 6 bins. As shown in Table \ref{tbl:SCT}, using orientation-discriminative features improves nearly all the performance metrics compared to using averaged features in appearance-based tracking. This can be attributed to the increased representation power leveraged by the orientation-based bins than the feature-average. Incorporating position likelihood into data association, as shown in  (\ref{eqn: pos_like}), significantly improves tracking performance, since it enforces spatial constraints relating to track-motion. Using orientation-discriminative features with position leads to small improvement over the average gallery in most of the performance metrics. 


\subsection{Multi-sensor Tracking}
\label{subsec:multi-sensor}
The DukeMTMC dataset is used for evaluating the multi-sensor tracking approach. This dataset consists of eight cameras placed on a university campus, with different orientations and heights. The proposed tracking approach is tested on videos from three cameras $2$, $5$, and $7$. The camera choice is based on the observation that they contain several people moving from one sensor FOV to another. Camera 2 and 5 collects high angle images from above while camera $7$ is situated close to the ground level, which closely simulates the view of sensors installed on ground robots. The testing is conducted on the training dataset of DukeMTMC for 4000 frames in each camera since the test set is not publicly available. Our detections are derived from Mask R-CNN, pre-trained on COCO \cite{lin2014microsoft} dataset, with a threshold confidence level of 0.91. The multi-camera performance is measured on commonly used IDP, IDR and IDF1 scores. 

The first step involves single camera tracking, which is performed by leveraging both, re-ID and position information. The re-ID features are computed from DaRe, trained on Market-1501 dataset.  For cross-sensor data association, we have utilized the $S2T$ ratio to divide the re-ID features corresponding to each track within a sensor into 6 bins. For three sensors, we perform three sets of cross-sensor data association i.e. between camera-2-to-camera-5, camera-5-to-camera-7, and camera-7-to-camera-2. For any given pair of two cameras, the data association is conducted as described in Section \ref{sec:multi_sensor_tracking}.
Lastly, unlikely track associations are dropped based upon a hand-tuned threshold.


We compare our results with a baseline approach, BIPCC \cite{ristani2016performance}, that is provided along with the dataset. BIPCC uses both re-ID features and position for multi-target tracking. Identity association is achieved by running a Binary Integer Program (BIP) to solve the correlation clustering problem on a weighted graph, where graph edges encode correlation between detections. 

Table \ref{tbl:MST} presents the multi-camera tracking results for both approaches. It is evident that our orientation-discriminative track-association approach outperforms the baseline by a large margin. This is a very interesting result because despite the disadvantage of our re-ID model, which is trained on a different dataset, the proposed method achieves better tracking performance than BIPCC, which is trained on the DukeMTMC dataset. This happens because BIPCC achieves cross-camera identity association by comparing the appearance within a temporal window, which contains a small variety of human poses, while our proposed orientation-discriminative representation maintains features of different poses in a person's whole trajectory. This enhanced representativenss leads to more accurate identity association across cameras.

\begin{table}[h]
\centering
\caption{Multi-sensor Tracking Performance on DukeMTMC dataset}
\begin{tabular}{|l|l|l|l|}
\hline
 \textbf{Metrics} & \textbf{IDF1} & \textbf{IDP} & \textbf{IDR} \\ \hline
 Baseline & 78.64 & 83.73 & 74.13 \\ \hline
  Ours & \textbf{82.78} & \textbf{84.79} & \textbf{80.86}  \\ \hline
\end{tabular}
\label{tbl:MST}
\end{table}

\section{Conclusion}
\label{sec:conclusion}
In conclusion, the proposed $S2T$ ratio helps in codifying the orientation information of a person in the image plane. Categorizing re-ID features into bins, based upon the $S2T$ ratio, significantly reduces the gallery size while retaining the semantic information associated with the features and leads to better re-ID performance. Further, our tracking results show that such a communication-efficient binned representation of the feature gallery improves D-MTMST performance.



The future work will extend the proposed approach to tracking in 3-D world frame and consider sensor egomotion. Realistic communication network conditions, such as intermittent or lossy transmission, will also be investigated.








\bibliographystyle{IEEEtran} 
\bibliography{egbib} 

\end{document}